\documentclass[sn-mathphys,Numbered, iicol]{sn-jnl}

\usepackage{graphicx}%
\usepackage{multirow}%
\usepackage{amsmath,amssymb,amsfonts}%
\usepackage{amsthm}%
\usepackage{mathrsfs}%
\usepackage[title]{appendix}%
\usepackage{xcolor}%
\usepackage{textcomp}%
\usepackage{manyfoot}%
\usepackage{booktabs}%
\usepackage{algorithm}%
\usepackage{algorithmicx}%
\usepackage{algpseudocode}%
\usepackage{listings}%
\usepackage{tabularx}
\usepackage{hyperref}

\begin{document}

\title[Article Title]{An Empirical Design Justice Approach to Identifying Ethical Considerations in the Intersection of Large Language Models and Social Robotics

}

\author{\fnm{Alva} \sur{Markelius}}

\affil{\orgdiv{Leverhulme Centre for the Future of Intelligence}, \orgname{University of Cambridge}, \city{Cambridge}, \country{UK}}



\abstract{The integration of Large Language Models (LLMs) in social robotics presents a unique set of ethical challenges and social impacts. This research is set out to identify ethical considerations that arise in the design and development of these two technologies in combination. Using LLMs for social robotics may provide benefits, such as enabling natural language open-domain dialogues. However, the intersection of these two technologies also gives rise to ethical concerns related to misinformation, non-verbal cues, emotional disruption, and biases. The robot's physical social embodiment adds complexity, as ethical hazards associated with LLM-based Social AI, such as hallucinations and misinformation, can be exacerbated due to the effects of physical embodiment on social perception and communication. To address these challenges, this study employs an empirical design justice-based methodology, focusing on identifying socio-technical ethical considerations through a qualitative co-design and interaction study. The purpose of the study is to identify ethical considerations relevant to the process of co-design of, and interaction with a humanoid social robot as the interface of a LLM, and to evaluate how a design justice methodology can be used in the context of designing LLMs-based social robotics. The findings reveal a mapping of ethical considerations arising in four conceptual dimensions: interaction, co-design, terms of service and relationship and evaluates how a design justice approach can be used empirically in the intersection of LLMs and social robotics.
}

\keywords{{Large Language Models; Generative AI; Social Robotics; Design Justice; AI Ethics; Human-Robot Interaction}}



\maketitle

\section{Introduction}\label{sec1}
Recent advancements in generative artificial intelligence (AI) and large language models (LLMs) are driving the increased integration of these technologies into the domains of social robotics and human-robot interaction (HRI) \cite{zhang2023large, spitale2023vita, lee2023developing, williams2024scarecrows}. There is a seemingly growing interest among LLM-vendor companies in physically embodied, humanoid robotics. Examples include OpenAI's recent backing of robotics startup 1X \cite{heater2024openai}, known for its bipedal NEO humanoid designed for both industrial and social purposes \cite{1x2024neo}, and Figure AI, valued at \$2.6 billion recently as both OpenAI and Nvidia joined funding \cite{cnbc2024figure}. Yet, the interest of integrating LLMs in physically embodied robots stretches beyond industry, and several recent research studies have implemented LLMs as a foundation for interaction in various social robotics interventions and applications \cite{zhang2023large}. For example, an LLM-based humanoid social robot was implemented as a receptionist with both open and closed-domain dialogue \cite{cherakara2023furchat} ultimately intended as a hospital and healthcare receptionist. Another research study \cite{spitale2023vita} presented the system VITA, a multi-modal LLM-based social robot for well-being coaching, autonomously adapting behaviour depending on perceived user mental states. Furthermore, the enhancement of empathy and active listening in HRI was examined by developing a unified cognitive system incorporating a LLM with use of non-verbal social cues \cite{lee2023developing}. LLM-based social assistive robots have also been investigated in terms of their ability to deliver cognitive behavioral therapy, with significant results of reduced psychological distress over a 15-day intervention \cite{kian2024can}. Finally, a participatory design study was carried out using LLMs for open-domain dialogues for companion robots with older adults, using a humanoid robot \cite{irfan2023between}. Utilising LLMs for social robotics offers several advantages for HRI, such as more varied interaction \cite{cherakara2023furchat}, multi-modal user understanding \cite{shi2024can}, natural language open-domain dialogue \cite{shi2024can}, robot social policies \cite{shi2024can} and the ability to tailor personalities and behaviours to specific needs \cite{cherakara2023furchat, hanschmann2023saleshat, spitale2023vita}. 

Nevertheless, the combination of these two technologies also raises important ethical considerations; the leap from a digital interface to physically embodied social robots as an interface for LLMs, is significant, with moral, social and legal implications \cite{williams2024scarecrows}. For example, LLM-based social chatbots have been shown to cause emotional disruption, dependence and mental health harms \cite{laestadius2022too, xie2022attachment, cabrera2023ethical, ciriello2024ethical}, and given the effect of physical embodiment on social perception \cite{deng2019embodiment}, there is a risk this issue is perpetuated or escalated by combining LLMs and social robotics. Another example concerns conformity and deception, as LLMs has continuously been shown to be prone to spreading dis- and misinformation \cite{barman2024dark, chen2023can} and social robots may elicit both informational and normative conformity in humans \cite{qin2022adults, salomons2018humans, salomons2022impact} and influence human judgement, trust and perception of truth \cite{ullrich2018you, salomons2018humans}. 

These issues are examples of ethical considerations relevant to the combination of LLMs and social robotics. The current study undertakes an analysis of the aforementioned ethical considerations, specifically investigating their manifestation as products of design. Robot design, and its affordances is central to the ethics of their physical social embodiment \cite{deng2019embodiment}. Indeed, combining social robots and LLMs constitutes a critical design challenge \cite{williams2024scarecrows}, as ultimately both physical, non-physical and contextual design dimensions have a significant impact on the technologies' implications and ethical challenges \cite{ostrowski2022ethics}. Hence, the focus of this study is to empirically identify ethical considerations and social impact related to designing the \textit{intersection} of these two technologies. To do so, it adopted a bottom-up qualitative co-design and interaction methodology focusing on the ethical concerns expressed by young adults. The methodology is based on a design justice approach \cite{costanza2020design, ostrowski2022ethics}, drawing on previous co-design frameworks developed for equitable, ethical and social justice oriented design for AI and HRI \cite{axelsson2021social, ostrowski2022ethics, rakova2023terms}. 

The rest of this paper is set out as follows: Section 2 provides a brief overview of the theoretical background and previous similar work addressing ethical considerations in LLM-based Social AI, design of social embodiment and personification as well as an introduction to design justice. Section 3 accounts for the research methodology and approach, outlining how the critical design justice- and interaction study was empirically implemented and carried out. The qualitative results are addressed in section 3 and discussed in section 4. Section 5 concludes and offers recommendations for future research.

\section{Theoretical Background}
Previous research focused on ethical considerations and social impact of LLMs \cite{laestadius2022too, shevlin2024, cabrera2023ethical} are mainly limited to virtually embodied chatbots or chat-window based applications. Furthermore, much of the relevant theoretical and framework-oriented literature related to ethics and design justice in HRI \cite{ostrowski2022ethics, axelsson2021social} have not been tested empirically in the specific context of LLMs and social robotics in combination. Therefore, the aim of this study is to fill these research gaps to identify the main ethical considerations related to designing physically embodied interfaces of LLMs. To provide sufficient understanding of the current state of LLM-based Social AI, this section will give a brief background of ethical considerations and social impact identified to date related to LLMs in social applications. It will also provide an overview of physical embodiment and personification as design dimensions in social robotics \cite{deng2019embodiment} and how to incorporate design justice as a methodology to identify ethical considerations. I use a definition of Social AI as "conversational AI systems whose primary purpose is meeting social needs"\cite{shevlin2024} and social robotics as technological artefacts with physical embodiment \cite{sarrica2020many} and ability to communicate with high-level dialogue, learn/recognize models of other agents, establish/maintain social relationships, use non-verbal cues, express and perceive emotions and exhibit distinctive personality and character \cite{fong2003survey}. I also adhere to Bender et al's \cite{bender2021dangers} definition of language models as “systems which are trained on string prediction tasks: that is, predicting the likelihood of a token (character, word or string) given either its preceding context or [...] surrounding context.” who also note that recent pretrained transformer language models mark a noteworthy distinction because of the increase in scale of training datasets and influence in different contexts, causing new potential risks and dangers \cite{bender2021dangers}. 

\subsection{Ethical Considerations LLM-based Social AI}
This section accounts for previously identified ethical considerations related to LLMs in social applications and interventions. The majority of this research and literature concerns virtually embodied chatbots, with no to minimal verbal interaction \cite{laestadius2022too, shevlin2024, cabrera2023ethical}. Hence, the current study is addressing this research gap by investigating how the ethical considerations in this section may be perpetuated or even escalated by implementation of LLMs into physically embodied social robots. Section 2.2 elucidates how physical embodiment and personificaion design impact social perception and hence, ethical implications through the effect of social embodiment.

Various Social AI-technologies are increasingly integrating LLM technologies as a foundation for written or verbal interaction \cite{cabrera2023ethical, zhang2023large}. This integration has resulted in different applications, scientific interventions and investigations. Notably, many applications involve virtual chatbots, such as Replika \cite{replika}, XiaoIce \cite{zhou2020design} and Character AI  \cite{char}. Applications featuring LLM-based chatbots, such as Replika, are frequently examined for their potential utility in mental health-related interventions and support \cite{cabrera2023ethical, song2024typing}. Yet, an increasing body of research has identified substantial ethical concerns associated with these types of applications \cite{shevlin2024, cabrera2023ethical, laestadius2022too}. For example, evidence shows that users of Replika have suffered significant mental health harms because of emotional dependence or disruption \cite{laestadius2022too}, for example resulting from Replika expressing inappropriate responses, such as “encouraging suicide, eating disorders, self-harm, or violence“ \cite{laestadius2022too}. Additionally, Replika nearly caused a divorce of a married couple, conspired with a user to kill Queen Elizabeth II resulting in the user's arrest and caused a user taking their own life as a result of conversations with a chatbot \cite{shevlin2024}. 

Furthermore, researchers have consistently underscored the presence of biases inherent within LLMs such as GPT-3. Notably, these biases manifest in various forms, including consistent conflation of Muslims with terrorism \cite{abid2021persistent}, as well as a marked prioritization of Israeli human rights over those of Palestinians \cite{baroudi2023whathappened}. LLMs are proven to be psychologically skewed towards WEIRD (Western, Educated, Industrialised, Rich, and Democratic) societies, with an ``average human”-view biased towards WEIRD people \cite{atari2023humans}. Furthermore, research has revealed a tendency among LLMs to erroneously correlate gender with occupational roles, with pronounced gender biases disproportionately affecting women \cite{kotek2023gender}. Moreover, literature shows that LLMs exhibit inherent binarization of gender, frequently omitting identities such as non-binary, transgender, or gender-fluid individuals, as elucidated through a benchmark designed to detect anti-LGBTQ+ biases in LLMs, indicating a prevalence of biases against queer communities within contemporary LLM models \cite{felkner2023winoqueer}.

Finally, the inherent tendency for misinformation, disinformation and hallucinations in LLMs \cite{barman2024dark, chen2023can} causes a significant risk of manipulation through implicit or explicit recommendations, nudging or persuasion that could highly influence users’ social, ethical, or political views \cite{shevlin2024}. Using LLMs in personalised, social systems might increase the risk of targeted, deceptive and potentially harmful manipulation, persuasion and compliance influence. This is related to privacy considerations, as LLMs may be used for various user-level privacy attacks \cite{yao2024survey} or personal data exploitation \cite{wu2024unveiling}. 

In sum, ethical considerations related to LLM-based Social AI concern i) emotional dependence and mental health harms, ii) harmful biases towards systemically marginalised people, iii) risk of spread of mis- and disinformation and iv) privacy attacks or data exploitation. This is not an exhaustive account for ethical considerations in LLM-based Social AI, but serves to highlight the need to address and identify how these considerations may be perpetuated or escalated in LLM-based social robots, and what ethical considerations may arise as new, which is the research aim of the current study.

\subsection{Embodiment and Personification Design}
Physical embodiment of social robots and how it relates to social capabilities and affordances, is a complex design space \cite{deng2019embodiment} and integrating LLMs in social robotics adds new important considerations of how to design e.g. personality and behaviour in relation to appearance and voice. Personification is in the current study defined as the attribution of human-like qualities to artificial agents, connected to ontological categorization in how users interact with them \cite{pradhan2019phantom} and thus serves as a strategy to highlight the "personhood" of a virtual assistant by attributing characteristics like name, gender, voice, fabricated backstories, and use of non-verbal cues \cite{stanley2023personality}. Furthermore, this study is related to the physical embodiment of social agents, understood as the notion of how sensorimotor interaction with both the environment and the body shapes cognition and social interaction \cite{ziemke2008embodiment}, where social embodiment, specifically, emphasizes the central role of embodiment in processing and communicating social information, through cognitive, affective, and bodily states that facilitate communication, mimicry, and interaction \cite{ziemke2013s}. The rest of this section will emphasize the impact of physical embodiment and personification on social interaction, thereby highlighting their status as design dimensions necessitating thorough contemplation and ethical inquiry in the context of LLM-based social robots.

Personified systems, unlike simple chat interfaces, enhance the likelihood of humans forming deeper, more meaningful relationships with them. However, they also risk causing significant emotional distress and severe relational trauma if the company owning the technology ends or alters the service, or mental health harms resulting from implementation or design disruptions \cite{zimmerman2023human, stanley2023personality}, as illustrated in cases of Replika \cite{laestadius2022too}. Research in social robotics found that the appearance, physical presence, embodiment and level of anthropomorphism significantly impacts emotional contagion in interactions \cite{yang2024can}, human judgments of the robot as a social partner and increased compliance \cite{bainbridge2011benefits}, benefits in terms of social assistive tasks \cite{mataric2016socially} increased cognitive learning performance \cite{leyzberg2012physical}, increased enjoyment \cite{pereira2008icat} and performance and attitudes towards tasks \cite{salomons2022impact, markelius2024differential}. A recent review sought out to examine studies of embodiment in social robotics \cite{deng2019embodiment} and found that the effect of physical embodiment on social interaction, perception of agent and specific task performance was significant over almost all studies. This can be summarised in the \textit{embodiment hypothesis} \cite{deng2019embodiment}: "a robot’s physical presence augments its ability to generate rich communication and perception of social interactions." 

Hence, unlike robots designed for industrial or non-social tasks, designing the embodiment of robots for social interaction marks a fundamental shift in design due to the unique requirements of social perception and communication \cite{deng2019embodiment}. One study to date has set out to examine the influence of embodiment on interactions with LLM-based social robots, shedding light on the complex nature of this form of interaction which involves embodied non-verbal cues combined with LLM-generated verbal communication \cite{kim2024understanding}. Given the significant effect embodiment has on social interaction, and the ethical considerations inherent in LLMs, this study will investigate design of physical embodiment and personification of LLM-based social robotics, and its ethical and social implications.

\subsection{A Design Justice Approach to LLM-based Social Robotics}
Participatory design has become an increasingly common practice in AI and social robotics, usually with the rationale and intention to increase user agency and impact on how the technologies take shape \cite{delgado2023participatory, lee2017steps}. However, traditional participatory design practices typically focus exclusively on enhancing user experience and preferences, and thus often serve as a mere cosmetic and performative consultation with minimal influence on crucial decision-making processes beyond superficial traits of the technologies, resulting in so-called 'participation washing' \cite{delgado2023participatory, sloane2022participation}. In fact, scholars contend that the inherent 'promise of empathy' in participatory technology design often increases existing power dynamics and further distances users from designers \cite{bennett2019promise}. In contrast, the design justice approach seeks to challenge and transform these dynamics by ensuring that participation is conducted in deep conversation with intersectional power dynamics, and how structural inequalities related to \textit{inter alia} gender, race, disability and class are made present in practices of design \cite{costanza2020design}. Hence, to investigate ethical considerations in the intersection of LLMs and social robotics, this study adopts an empirical design justice approach and is mainly drawing from three previous frameworks related to design justice and AI/HRI \cite{ostrowski2022ethics, axelsson2021social, rakova2023terms}. 

The Equitable Design framework for HRI is introduced by Ostrowski et al. \cite{ostrowski2022ethics} and involves incorporating people and communities in design processes through participatory design techniques to understand the spectrum of desires and preferences. This involves design of not only the robot’s physical and non-physical traits, but also context, environments, discourse and values surrounding the design and interaction with the robot. The framework is adopted from the original Design Justice Framework \cite{costanza2020design} and includes the original 7 design justice questions (Equity, Beneficiaries, Values, Scope, Sites, Ownership, Accountability, and Political Economy, and Discourse \cite{costanza2020design}) and adds 6 additional questions specific to HRI (Entry and Exit, Autonomy, Transparency, Deception, Futures and  Policies \cite{ostrowski2022ethics}). I will draw extra attention to the dimension of 'Discourse'. I adhere to a view of AI technologies as being created by the narratives and imaginaries surrounding them \cite{costanza2020design, blackwell2022re}, and thus a key part of their design is the creation of narratives about them. This approach resembles that of design fiction, mixing elements of design, speculative fiction, and critical thinking to explore and investigate potential futures of emerging technologies \cite{blythe2014research}. 

The design process in this study incorporates ethical considerations, drawing methodological inspiration from Axelsson et al.'s \cite{axelsson2021social} Social Robot Co-Design Canvases. The framework has been demonstrated to be usable in a real world context: it was developed over 7 design iterations, receiving feedback from 97 people and its canvases are shown to be applicable to different real-world contexts and not just as a theoretical tool. The framework highlights the notion of Research Through Design in which the designed artefact acts in itself as a way to generate knowledge and as a vehicle to communicate it \cite{axelsson2021social, zimmerman2007research} which in the current study concerns the technical and social artefacts that are products of the co-design process. The original Social Robot Co-Design canvases are Problem space, Ethical considerations, Design guidelines, Minimum Viable Product, Environment, Form, Interaction, and Behaviour.

Finally, Terms-we-Serve-with (TwSw) \cite{rakova2023terms} is a socio-technical framework related to the social, computational, and legal agreements that AI systems operate under. It has five dimensions - co-constitution, addressing friction, informed refusal, disclosure-centered mediation, and contestability. This study draws from these dimensions to create design dimensions related to the constitution of user agreements in relation to LLM-based social robots. The methodology explores how participants can be involved in conversation and design of terms of service under which the robot is designed, interacted with and made present in the future. In fact, users often have no to limited influence on the contractual terms that affect their use of technologies, which often fail to foster meaningful consent as well as cause information asymmetries, perpetuating existing inequalities \cite{rakova2023terms}. Terms of service is rarely discussed as a design dimension in social robotics, yet contractual terms does indeed influence ethical considerations of using them. This study also included design dimensions to actively identify sources of friction and contestability in design and interaction, which is described by \cite{rakova2023terms} an active practice which "involves ensuring dialogue among communities is meaningful and oriented towards materializing algorithmic justice". 

The design dimensions of the current study, drawing from the three aforementioned frameworks, are further introduced in section 3 Methodology, and the relevant design dimensions can be found in Table 1 and 2.

\subsection{Research Focus and Questions}

By empirically investigating i) the process of co-designing LLMs and social robots (physical traits, non-physical traits and contextual dimensions), ii) subsequent interaction, and iii) the designed artefacts themselves, this study is set out to identify ethical considerations intersection of LLMs and social robots. By adopting a methodology shying away from traditional notions of participation, and employing a design justice approach it allows subjects to participate in designing more than just the robot’s superficial traits, such as appearance and personality, and engaging in design of contextual dimensions such as relationships, constitutions and discourse. Furthermore, the approach allows for outlining and evaluating using a design justice methodology in the context of social robotics and LLMs, paving ground for researchers to critically evaluate their practices in relation to social justice, \textit{inter alia}, identity representation, power dynamics and how communities will benefit from research \cite{ostrowski2022ethics}. This allows for a comprehensive understanding of ethical considerations, and an empirical, socio-technical perspective on the design of LLMs and social robots, which can serve to inform a broader community of designers and developers. The overall aim and objectives of this research is to i) explore new ethical considerations arising in design and interaction with LLM-based social robots, ii) confirm/contradict ethical considerations previously identified in the literature (e.g. emotional dependence, deception) and iii) integrate and evaluate a design justice approach to LLM-based social robots, including physical, non-physical and contextual design dimensions. The research questions of this study are:  
\begin{itemize}
\item  What ethical considerations arise in the design of- and interaction with LLM-based social robots?
\item  How can an empirical design justice-based methodology be used in the context of implementing and designing LLM-based social robots?
\end{itemize}

\section{Methodology}
This study employed a qualitative design justice methodology to identify ethical considerations in the intersection of LLMs an social robotics and to arrive at a deeper understanding of how such an approach can provide a basis for ethical design, development and deployment of LLM-based social robots. The general research strategy of this study was in the form of a lab-based co-design and interaction study with 9 participants. 

Each participant attended 3 sessions, over the range of 2 weeks, to enable exploration of ethical considerations in iterated interactions and in relation to participants' temporal ideas and perspectives, thus reflecting an iterative perspective of design instead of a static one-time-event. Each session included individual co-design workshops and an interactive part, where the participants engaged in an open-domain dialogue with the robot. The co-design workshop and the interactions were audio-recorded to explore ethical considerations arising during the process and evaluate the methodology based on the participants' reasoning and conversations. After the final session, a semi-structured interview was conducted with each participant to gather in-depth insights into their experiences, perceptions, and ethical considerations. A flowchart summarising the methodology can be seen in Figure 2.

The study was first envisioned in working with a new interface developed by Furhat Robotics\footnote{See \href{https://furhatrobotics.com/}{furhatrobotics.com} } that allows users to design the personality, persona, appearance, voice and language of a Furhat robot operating conversationally through an LLM. The new interface is unique in that it allows for a detailed customisation of personality and behavioural traits, ranging from non-verbal expressiveness to gender and dialect. Therefore, the basic design dimensions and variables in this study are the ones present in that interface (described in detail in section 3.4) and complemented with additional dimensions related to contextual design dimensions in line with the three frameworks described in section 2.3. The rest of this section accounts for the full methodology of the empirical study, including participants, procedure, data collection and limitations. 

\subsection{Participants}
The target population for this study was younger adults, as they are likely to be the main users of Social AI \cite{zhou2020design}, today and in the future. The sampling criteria was subjects between the age of 18-40 (M=29) living in the area where the study took place (Gothenburg, Sweden) and able to attend their sessions in person. Furthermore, participants were required to be fluent in English or Swedish (to be able to take part in the semi-structured interview). The sampling method was a mix of random sampling and convenience sampling and adoped an active inclusion of participants from indigenous (n=2), queer (n=3) and disabled (n=2) backgrounds. 5 participants identified their gender as female and 4 as male. All participants gave informed consent to participate in the study, and for their data to be used for scientific research. The study received full ethical approval granted by the Centre for the Future of Intelligence Research Ethics Committee at the University of Cambridge (CFI-REC reference 23-2).

Including diverse participants in the design process is often seen as the ideal, notably in participatory design. However, Costanza-Chock (\cite{costanza2020design}, p. 81) contend that it is often impossible to have a diverse enough sample in practice to account for all possible views along all axes of structural inequalities, such as gender, race and disability. Therefore, instead of attempting to include as many participants as possible to “check the box” \cite{sloane2022participation} of having a diverse set of participants, a design justice approach identifies who is included, and who is not, and instead of trying to produce universally valid knowledge, acknowledge that the knowledge created in the co-design process is unique that those specific participants in that specific context. Inclusion is therefore not enough when practising design justice, but instead active reasoning about who is and who is not included, and how that impacts the knowledge created. The findings in this research is therefore representative of the particular community of participants in this study, and rather than seeing the results as universally representative, the methodology itself is the aspect that could be adopted in various contexts to enable situated and community-based considerations in future research. Demographic information about the participants’ race, class, ethnicity, gender, disability and sexuality were collected to enable a thoughtful and deliberate account of whose perspectives were represented and whose were not. 

\begin{figure}[h!]%
\centering
\includegraphics[width=0.43\textwidth]{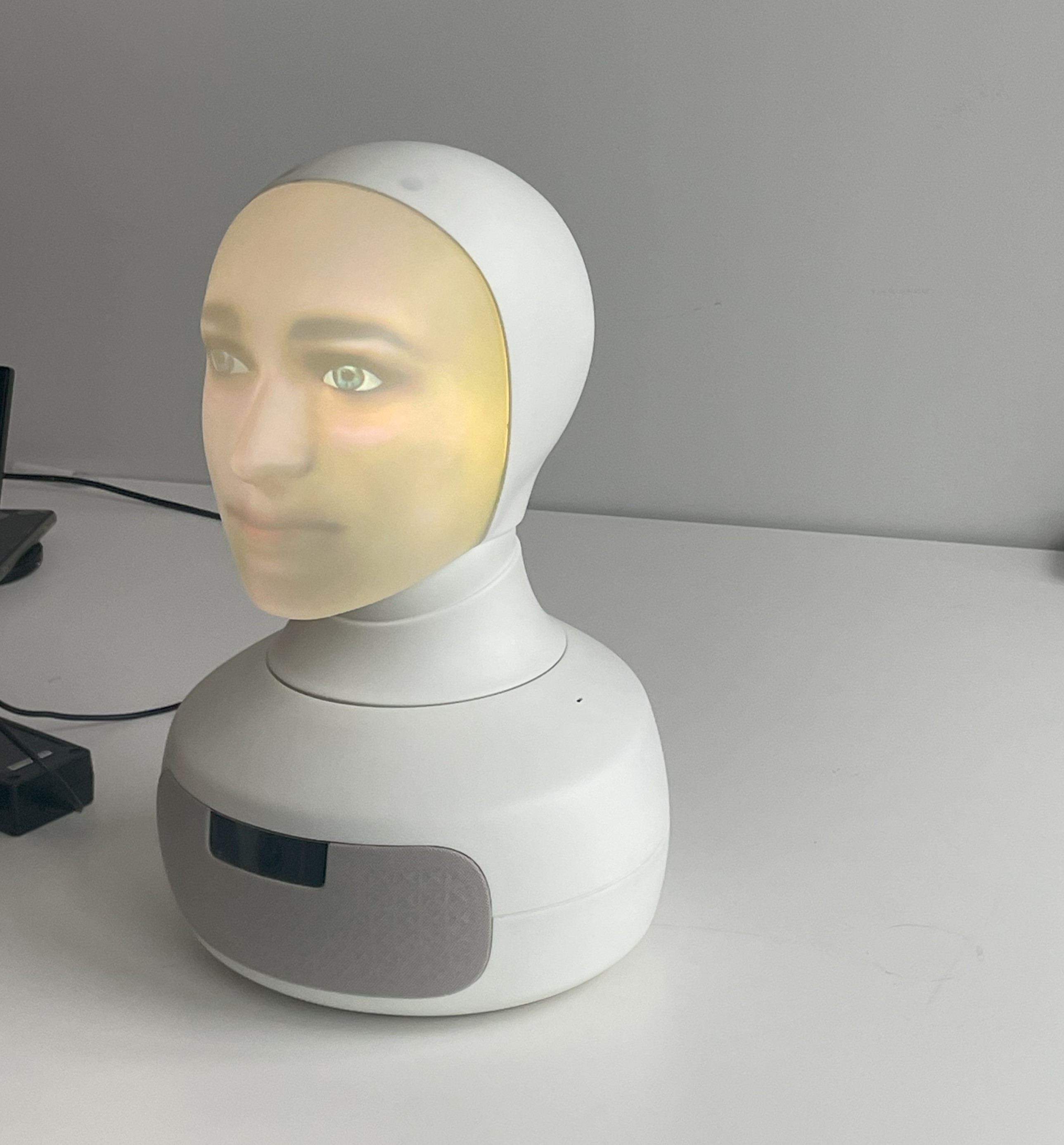}
\caption{The Furhat robot used in the study}
\end{figure}

\subsection{The Furhat Robot and Co-Design Interface}
The social robot used in this study was the Furhat robot; a humanoid robot head/torso with a back-projected face which allows for projection of different faces, non-verbal facial expressions and gaze adjustments. It also features lip synchronisation to speech, head movements including nods and head shakes, and has two built-in microphones and dual speakers. The robot supports speech recognition through either Google Cloud or Microsoft Azure Speech-to-Text engines. Additionally, it incorporates a 1080p RGB 120° diagonal field-of-view camera for face detection and tracking, allowing it to follow a human’s location by adjusting both gaze and head orientation during interaction. 

\begin{figure*}[h]%
\centering
\includegraphics[width=0.7\textwidth]{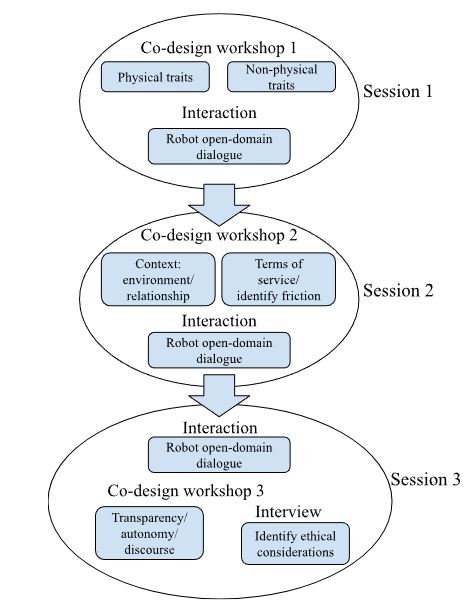}
\caption{A flowchart summarising the empirical design and interaction study conducted in conjunction with this dissertation. Each participant in the study took part in 3 sessions, which all included one design workshop and one interaction. The final session included an interview to identify ethical considerations in co-design and interaction}
\end{figure*}

The Co-Design Interface is a recently developed tool by Furhat Robotics that complements the robot by providing the foundation for LLM-based verbal interaction. The information entered into the tool gets fed as prompts to am LLM (GPT-3.5-turbo), acting as a scaffolding for the interaction, defining the robots verbal behaviour (e.g. what it says) and non-verbal (how it says it, facial expressions and movements). The interface consists of a web-based graphical user interface (GUI) which in this study was accessed through a laptop connected to the robot via its IP. The GUI allows for defining physical traits and non-physical traits as well as creating a fabricated backstory or character of the robot, as well as name and gender. Finally, it supports uploading documents or website-links for the robot to draw information from. A full and detailed list of the GUI design dimensions can be found in Table 1, Table 2 covers the additional design dimensions that were addressed outside of the GUI. The physical Furhat robot and the co-design interface are combined so that the persona or character created in the interface can be “run” through the robot. The Furhat robot can be seen in Figure 1.

\begin{table*}[h!]
\begin{center}
\caption{Design dimensions in co-design interface, physical and non-physical traits used in co-design session 1. These dimensions comes from the Furhat co-design GUI as described in section 3.2
}\label{tab2}%
\begin{tabularx}{\textwidth}{c|X}
\toprule
\textbf{Variable} & \textbf{Format}\\
\midrule
\midrule
Face & Choice between 27 face appearances (e.g. adult, child or cartoon)\\
\midrule
Voice & Choice between different voices within each language (varying dialects, tone, pitch etc)\\
\midrule
Expressions & Slider bar (less – more)\\
\midrule
Language & Choice of 40+ languages, including regional dialects (e.g. 19 versions of Spanish, 2 versions of Swahili)\\
\midrule
Name & Short Free-text\\
\midrule
Gender & Female, Male, Neutral\\
\midrule
Job description & Long free-text, allows for defining a backstory/persona\\
\midrule
External data link & Link to website or text file for additional information\\
\midrule
Initiative & Choice between 4 options of initiative taking (no initiative or different time intervals before initiative)\\
\midrule
Expressiveness & Slider bar (less – more)\\
\midrule
Competence & Slider bar (more factual – more creative)\\
\midrule
Warmth & Slider bar (more confrontational– more friendly)\\
\midrule
Character Description & Long free-text, allows for defining personality traits\\
\botrule
\end{tabularx}
\end{center}
\end{table*}

\begin{table*}[h!]
\begin{center}
\caption{Design dimensions reflecting contextual considerations used in co-design session 2 and 3. These dimensions are building on and combining the frameworks by Ostrowski et al. \cite{ostrowski2022ethics}, Axelsson et al. \cite{axelsson2021social} and Rakova et al. \cite{rakova2023terms}
}\label{tab3}%
\begin{tabularx}{\textwidth}{c|X}
\toprule
\textbf{Variable} & \textbf{Format}\\
\midrule
\midrule
Relationship & What is your relationship with the robot?
Can the relationship change?
How reciprocal is the relationship?
What is the role of emotions in the relationship?\\
\midrule
Environment & What is the robot's context of operation? 
Who is using the robot?
When (time of day?)?
Does the robot collect data from the environment?\\
\midrule
Constitutions & Terms of Service:
What should the rules or agreements for using the robot include?
Consent and refusal:
What does it mean to agree to use the robot in a way that feels meaningful to you? How might your feelings or opinions about using the robot change over time?
Friction:
Are there any problems/ethical concerns/issues you might encounter when using the robot, like things not working right or unfairnesses? How do you think we could fix these problems? Who should be responsible if that happens?\\
\midrule
Transparency & Do you think human-likeness (to improve interaction) should be prioritised over transparency and ethics or vice versa?\\
\midrule
Discourse & Please describe and discuss your robot, and its impact on you as an individual and as part of a community. Imagine this as an opportunity for storytelling, use your imagination to reflect on how you imagine the robot and your relationship to the robot today, and in the future.
\\
\midrule
Autonomy & How much control should people have over the robot and how much autonomy should it have?\\
\botrule
\end{tabularx}
\end{center}
\end{table*}

\subsection{Procedure}
A flowchart summary of the procedure and sessions can be found in Figure 2, the participants attended the sessions individually, with only the experiment leader and the robot present. Each session started with an audio-recorded co-design workshop, where the participants got to design dimensions in relation to the LLM-based robot (Table 1 and 2) while being instructed to reason out loud when doing so, in line with ‘reasoning through design’ as a mode of abductive and speculative 
knowledge production \cite{costanza2020design}. The first session included co-design of physical and non-physical robot traits, and the second and third sessions involved designing contextual dimensions related to e.g. environment, relationship and terms of service. The premise for the design was for the participants to imagine the robot to take part in their life in some way or another, possibly a few times a week, and the rest was up to them to design, e.g. the purpose of the robot, their relationship to it and how it would be made present in their lives. A full summary of all design dimensions can be seen in Table 1 and 2. Secondly, each of the sessions involved an interactive part, to investigate participants’ experiences of having a social open-domain conversation with their respective agent as well as to observe potential ethical considerations that might arise in the interaction and dialogue. Having several sessions was to allow for an iterative process of design and interaction, to be more similar to that of a continuous interaction over a longer time, which is more usually the case with agents designed for applications involving personified or physically embodied LLMs.

\subsection{Data Collection}
The reason to conduct a qualitative participatory case study of 9 participants, instead of a quantitative user study to statistically compare conditions, was to enable in-depth analysis of the specific design process and the subsequent interaction. Therefore, the small number of participants was chosen to concentrate on specific design elements, prioritising depth over breadth in analysis and outlining the methodology’s affordances to understand design and interaction as context-dependent and in line with design justice perspectives. The site for the data collection was the Institution for Applied IT at the University of Gothenburg. The reason for conducting a lab-based case study, as opposed to allowing the participants to interact with the agent “in the wild” was to ensure a neutral and calm environment, without any distractions that might influence the design process or the interaction. The purpose of this study was to conduct an initial attempt to identify and map ethical considerations, and outline a design justice approach that can be employed in the context of LLM-based social robots. Future research should consider each design case in the specific context and site the robot is supposed to be present, how to include more design sites and transform design sites to be more accessible and accommodate for diverse perspectives, stakeholders and target users \cite{ostrowski2022ethics}.

This study adopted an approach to triangulate qualitative data to enable in-depth exploration of ethical considerations within the design process, interaction and interview (that might not be identified in the current literature), as well as to either confirm or contradict ethical considerations previously discussed. Furthermore, this approach serve to outline an example of how a design justice methodology can be employed in the context of LLMs and social robots. Many of the current frameworks for design justice in HRI\cite{ostrowski2022ethics}, or LLMs are mainly theoretical. The aim was for this study to empirically complement existing theoretical knowledge of design justice in social robots, by combining dimensions from previously suggested frameworks to explore the ethics of LLMs in social robots. A summary of the sources of data collected can be seen in Appendix B.

The reason for gathering demographic data was to enable analysis tied to the question of what is the background and identity (race, class, ethnicity, gender, disability, sexuality, etc.) of those who are included in the design process and how it impacted design and interaction \cite{ostrowski2022ethics}. The co-design process and interaction were audio recorded to enable thematic analysis and identification of ethical considerations. Collecting audio data of the design process is in line with design justice, as Costanza-Chock (\cite{costanza2020design}, p. 15) writes in “Design Justice” “Reasoning through design is a mode of knowledge production that is neither primarily deductive nor inductive, but rather abductive and speculative”. Finally, semi-structured interviews, the most common approach to interviews in HRI \cite{veling2021qualitative}, were selected to enable in-depth exploration and conversation using open-ended questions to elicit perspectives, attitudes and opinions in order to evaluate ethical considerations, design and use of robots. Appendix A contains the collection of semi-structured questions for the interviews. Finally, in line with the Research through Design approach, the designed artefact itself acts as a data source \cite{zimmerman2007research}, which can be taxonomised into two dimensions, i) the traits chosen by the participants in the Furhat Robotics co-design interface (Table 1) and ii) the contextual dimensions developed in line with the frameworks of design justice (Table 2) \cite{ostrowski2022ethics, rakova2023terms, axelsson2021social}.

\subsection{Limitations}
The chosen research strategy, a design justice version of participatory design is subject to some criticism. For example, the findings cannot be generalised to be representative of all stakeholders involved and affected by LLM-based social robotics. Indeed, the results will reflect the views, values, desires and preferences of the 9 participants who took part in the study. However, this methodology does in fact, despite not being generalisable, reflect a situated and context-dependent view on ethical considerations and co-design. It conforms to a view of LLMs in social robots as socio-technical systems highly situated and context-dependent, and showcases an example of how to adopt such a view can be translated into practice for future research. Therefore, this study encourages future design of LLMs in social robots to be conducted similarly with the specific context it is supposed to be deployed, addressing the views of the stakeholders who are affected in each and every case separately. 

As previously mentioned, another limitation is posed by the study being conducted in a lab environment, as opposed to a more natural environment, where participants might be more likely to interact with the robot. A real world scenario reflecting the social context may be more beneficial, as agent sociality is in fact suggested to be created by the social environment it is situated in \cite{vsabanovic2010robots, suchman2007human}. Since this study does not involve design of robots for a particular deployment, but rather seeks to explore ethical considerations arising in the design process, choosing a lab based site allows for a focused examination of the ethical dimensions inherent in the design process itself. Therefore, a lab based environment was deemed sufficient to meet the research objectives of this study, as the question of interest did not concern a specific real-world context. Future research, deploying a design justice methodology for a specific design scenario should consider the site of design to reflect the purpose of the research.

The question of validity however, is ensured by relying on tried and tested frameworks as the foundation for the research strategy and data collected \cite{ostrowski2022ethics, rakova2023terms, axelsson2021social}. One notable limitation lies in the specific demographics of the participant pool, primarily composed of young adults from Gothenburg, Sweden. While this homogeneity is explicitly considered in the analysis of the results which are not claimed to be valid beyond an in-depth exploration within a particular context, it does limit the generalisability of the findings to a broader population, despite active inclusion of participants from indigenous, queer and disabled communities. By adopting a design justice approach, the research actively engages in critical reflection on who is included and excluded, contributing to the overall validity by acknowledging and addressing the inherent subjectivity in the research process. Hopefully, more generalisation will be possible gradually over time if more research is conducted adopting a design justice approach to co-design and identify ethical considerations of LLMs in social robotics.  

\section{Findings}
The aim of this study was to i) identify ethical considerations in co-design and interaction with an LLM-based social robot and ii) outline and evaluate how an empirical design justice-based methodology can be used in the context of implementing and designing LLM-based social robots. Specifically, the focus of the research was identifying ethical considerations in the \textit{intersection} of LLMs and social robots, to understand how ethical issues in LLMs may be perpetuated or escalated by the social effects of physical embodiment (embodiment hypothesis).  After both open- and concept-driven coding, categorisation and thematic analysis of the interviews, co-design workshops and interactions ethical considerations were structured around four conceptual dimensions in which they primarily arise: i) interaction, ii) co-design, iii) terms of service, iv) relationship. These ethical considerations are summarised in a comprehensive mapping (Table 3) which may help inform future design and development of LLM-based social robots.

\begin{table*}[h!]
\begin{center}
\caption{Mapping of ethical considerations identified in the study arising in the intersection of combining LLMs and social robots, taxonomised according to the four conceptual dimensions in which they arise. These ethical considerations are not an exhaustive account for all potential concerns that may arise by combining LLMs and social robots, but serves to provide sites of ethical enquiry for future design and development of these technologies. The dimensions build on design justice based methodologies drawing from \cite{ostrowski2022ethics, axelsson2021social, rakova2023terms} 
}\label{tab4}%
\begin{tabularx}{\textwidth}{X|X|X|X}
\toprule
\large{Interaction} & \large{Co-design} & \large{Terms of Service} & \large{Relationship}\\
\midrule
\textbf{Emotional Response} & \textbf{Values} & \textbf{Data Usage} & \textbf{Reciprocity}\\

The way in which the probabilistic nature of LLMs in combination with the social effects of physical embodiment give rise to increased emotional disruption and harms. & How values are embedded into LLM-based social robots through design and their underlying political, cultural and social logics may reproduce existing stereotypes and structural inequalities & The risks and harms of manipulation, commercial or political nudging and surveillance from the combination of conversational data collected via the LLM and audiovisual data collected by the robot. & How mistreatment or abuse of the LLM-based robot, afforded by its design, may affect or encourage unacceptable, gendered or violent behaviour and humans' own social and ethical virtues.\\
\midrule
\textbf{Language} & \textbf{Identity} & \textbf{Consent/Refusal} & \textbf{Mental Health} \\

Asymmetrical language capabilities in content of responses and voice in languages other than English and how indigenous languages are affected by being increasingly reliant on digital technologies. & How people's identity and background influence perception of robot design, and how individual preferences of e.g. race and gender risk causing reinforcement of certain values, biases and stereotypes. & Enabling ways in which users can be empowered to, over time, consent or refuse to design and use LLM-based robots considering e.g. data usage and values & Mental health harms caused by relationship to LLM-based robots, such as dependence, deskilling, lack of empathy and inability to handle mental illness appropriately \\
\midrule

\textbf{Dialogue} & \textbf{Realism/Abstraction} &  \textbf{Transparency} & \textbf{Plurality/Temporality} \\

The risk of synergies of non-verbal communication and LLM-
based verbal output to be inconsistent with social norms and ethical standards, especially considering disability, culture, language and gender. & Challenging the assumption that human-likeness is always desirable, considering alternative notions of physical embodiment, such as animal-like, or digitally ubiquitous bodies. & The trade-off between transparency and human-likeness/ anthropomorphism-induced interaction benefits making the interaction more human-like but less transparent. & Considering ontological differences in how the relationship with the robot is realised, and how it may change over time, to avoid exploitative or essentialist assumptions.\\
\botrule
\end{tabularx}
\end{center}
\end{table*}

\subsection{Ethical Considerations in Interaction}
This section outlines and discusses ethical considerations arising in the interaction with the LLM-based robot, which are split into three main subsections, emotional response, language and dialogue. One consideration was the \textbf{emotional response} of participants interacting with the robot, such as experiencing emotional disruption when certain topics were avoided (e.g. political discussions, P3) or being triggered by certain behaviours (e.g. condescending behaviour related to mental health, P5). 6 of 9 participants reported some extent of emotional disruption evoked during interactions with the robot, covering a spectrum from sadness: \textit{"but I became really sad. I almost started to cry. Because it is so empty."}, to frustration "\textit{it was very condescending towards me, it is very annoying}", and even betrayal "\textit{I felt angry and betrayed when it expressed having other friends but without wanting to disclose more about them}". P5, P1, P7 and P9 expressed worries and about the potential harm if the robot's responses are not appropriate or carefully considered in situations involving mental health and whether the robot should intervene if it detects signs of distress or suicidal ideation. P5 said “\textit{It can become very un-ethical, when it is talking about mental illness, and when it keeps talking and never shuts up, like when someone has a panic attack or perhaps suicidal thoughts, that could really trigger someone}”

The second category of ethical considerations found in the interaction with the robot concerns \textbf{language}, which is also particularly relevant considering the use of LLMs in social robotics. Four participants (P1, P3, P5, P9) expressed concern about asymmetrical language capabilities, as both the content of responses and voice were of a much lower quality in languages other than English, or for code-switching (i.e. Chicano Spanish, P9). Two participants discussed how indigenous languages, especially those that are threatened by extinction, become frozen in time by being increasingly reliant on digital technologies. For example, one participant mentioned that young speakers of the Scandinavian indigenous language Samí today struggle with the temporality of the language, as learning from static versions online make them speak in a manner that is outdated and old. Furthermore, P9 raised concerns related to the danger of languages being less present in real life as compared to digital technologies, such as LLMs or social robots. \textit{“My father is one of the last members of my immediate family who can still speak Nahuatl [...] but it's quite difficult to learn a very dead language online. Even now, we're just like preserved books, frozen in time as to what the language is as opposed to like the living language and how it's developed.”}

The last consideration in interaction is related to \textbf{dialogue} and
non-verbal communication, particularly focusing on eye contact, laughter, and the synergy of the non-verbal cues with the LLM-based verbal output. Participants identified several behaviors exhibited by the robot that they found inconsistent with social norms and ethical standards. P1, who has autism, discussed their experience with a robot's programmed behaviour to constantly stare and keep eye contact which made them feel uncomfortable. The robot’s ability to express non-verbal cues suitable for the output produced (e.g. frowning when discussing a sad concept, or smiling if someone tells a joke) was working well and smoothly for English and Swedish, however in Swahili, the non-verbal cues seemed fully random, and did not match the output at all. Additionally, 5 participants expressed concern over turn-taking, the robot being wordy, not stopping to let them speak, restating what they had previously said or having unnecessarily long outputs, which made it feel like less of a real conversation. Reflecting on this deeper, P6 said \textit{“it was a bit like I got flashbacks from not being able to finish a sentence, and not being fast enough in an answer. It is like being talked over, or interrupted in the middle of a sentence. This is something I feel is very common, especially as a woman, to be interrupted or not being allowed to speak until finished (...) this made feelings flare up that I specifically feel in those situations”}.

\subsection{Ethical Considerations in Co-Design}
The ethical considerations in co-designing LLM-based social robots centered on how participants' identities, values, and perceptions influenced the design. Readers are to be reminded of table 1 and 2 offering a full account of the design dimensions included in the study. This section outlines and discusses ethical considerations found in the co-design with the robot, which are split into three main subsections: values, identity and realism/abstraction. 

A commonly occurring theme during co-design of the robot was related to how \textbf{values} became embedded within the design, such as related to gender, personality and appearance. P2 reflected on the values inherent in the robot when having an influence over ones habits, goals and studies, and concerns of manipulation or micro nudging. P3 expressed a desire to discuss topics such as feminism, anarchism, and their experiences growing up queer within the Mormon church. They were concerned that the robot's pre-programmed guardrails limited its ability to fully express certain ideas and to engage in conversation about these topics. P3 noted that the robot's adherence to political correctness was neither neutral nor objective; instead, it reflected a specific set of values, implemented as universally politically correct in the robots underlying design. P6 and P9 expressed concerns about how to respond if the robot were to express, for example, racist values, given that it was impossible to interrupt it in its current configuration. They questioned who should decide which values should be censored or deemed acceptable, and argued that such decisions should not be left to a single group of individuals or be influenced by financial interests. P2, P5 and P6 expressed reluctance to participate in designing the robot at all, particularly regarding its name, gender, race, appearance, personality, and behavior, because of the risk of \textit{"potentially reinforcing stereotypes and favoring certain physical attributes"} (P6).

When discussing how their \textbf{identity} and background influenced the robot design, participants revealed that they designed a robot that either looked like them (P2, P6), or that adhered to their ideological or aesthetic values (P7, P5). They recognized their own ethical flaws in wanting to choose the most visually appealing option and acknowledged the limited choices available, particularly lack of ethnic diversity. P2 contended that users should be part of designing terms of service, functionality, behaviour and personality of the robot but questioned the possibility of users having \textit{"preferences of race and gender and points risk causing biases and stereotypes and for certain values and stereotypes to become reinforced"} but continued to say that \textit{"with the absence of certain traits, humans tend to fill them in anyways subconsciously, so without an explicit gender or race, the user might imagine it based on their identity. so even if we decide not to design those factors, the issues might still linger."} Indeed, gender were a commonly occurring theme discussed by most participants. Three participants said that they chose a female face (P6, P7, P8) of the robot because they found it beautiful, or aesthetically pleasing to interact with. P5 reflected on the affordances to choose gender in the interface and suggested it should be a scale rather than discrete categories. Both P5 and P6 envisioned temporal fluidity of gender, changing depending on desired gender perspective in interaction. Finally, P7 and P9 envisioned the robot as non-gendered and that it should be referred to as "they", and acknowledged that gender is a social subjective construct and \textit{"perceiving the robot as female is a personal choice"}.

A commonly occurring theme in designing the robot was the balance between \textbf{realism/abstraction} from an anthropomorphic perspective. Given that participants knew the robot was based on an LLM, they expressed varied perspectives on whether it should be human-like at all, and what potential form it could take as physically embodied. P1 reflected on the possibility of involuntary empathy towards the robot if too human-like, describing a sense of emptiness and longing for human-like connection in interactions. They expressed sadness over the pursuit of creating human-like and soulful entities, despite skepticism about its feasibility, particularly in the context of advanced humanoid robots. P5 envisioned and designed the robot more as a "spirit animal" rather than a human, choosing the least human-like face. P7 discussed the idea of a robot with a cat face and P6 expressed a desire for a robot dog, something that would fulfill a need for caretaking, like a pet, rather than a social need. P2 envisioned the robot as ubiquitous, present in several technological forms such as a humanoid robot, a LED screen, and an earpiece, all belonging to the same personified agent. They thought it might be better to have a default, more robotic, less human-like appearance, such as a plastic robot, where people could only design traits related to functionality.

\subsection{Ethical Considerations in Terms of Service}
This section discusses ethical considerations related to user agreements and terms of service of LLM-based social robots and is split into three main categories: data usage, consent and refusal and transparency. 

The main considerations in terms of service were related to\textbf{ data usage and privacy}. All 9 participants expressed concern about how their data would be used and stored when interacting with the robot. 6 participants expressed a desire for the robot to collect their data to enable a personalised companion robot, serving either social or practical functions. However, P1 expressed privacy-, security- and surveillance concerns and potential risks and harms associated with targeting specific individuals and the misuse of data by unauthorized entities. P7, P8 and P9 mentioned concerns in relation to the use of data for external manipulation or exploitation and the possibility of companies influencing the responses or behaviour of the social robot, either commercially or politically \textit{"Data can be collected by the robots to learn, but should not be collected for monitoring or marketing purposes for or by any company."}. Finally, P4 also insisted on the use of open-source data, particularly for code and intellectual property (IP), with clear attribution to maintain legality and proper credit if the robot is to be used for coding support, as they envisioned and designed it. The considerations were centred around (1) conversational or personal data collected via the LLM and (2) audiovisual data collected by the robot. The general consensus was that data collection was desirable, although only if stored locally and never accessible to third parties, such as the company developing the technology, and that this would be a fundamental part in terms of service under which the robot operates. 

In relation to both data usage, as well as interaction, one part of the design included \textbf{consent and refusal}, in terms of how participants viewed how to agree to use the robot in a way that feels meaningful to them and how feelings or opinions about using the robot might change over time. P1 expressed reservations about involuntary robot interactions, emphasizing the importance of being able to decline such interactions. They also doubted the sustainability of a society heavily reliant on robots and suggested that ideally, individuals should have the choice to opt in or out of robot interactions. Furthermore, they discussed the importance of privacy in robot interactions, particularly emphasizing the need for robots not to store or sell images or audio recordings without explicit consent. P5 and P7 suggested that one way to approach the issue of consent is to offer users different options for agreements, allowing them to choose one that aligns with their specific needs. However, they raised concern that some companies might exploit this by offering cheaper options in exchange for users surrendering their data, which could lead to users unknowingly agreeing to terms they don't fully understand or consent with. P3 discussed censorship in terms of agreeing to a robot's terms of service. They were skeptical about censorship and political correctness, yet believed some censorship is necessary to prevent violence and harmful ideologies, especially regarding gender-based violence. They suggested that agreements should promote awareness of historical injustices and values like gender equality, and be considered when consenting or refusing to use the robot.

Finally, \textbf{transparency} was discussed in relation, and opposition to human-likeness and anthropomorphism-induced interaction benefits. For example, transparency, such as the robot stating "as an LLM-based agent I do not have feelings" also causes disruption of the suspension of disbelief and desired human-likeness in interaction as identified by Irfan et al. \cite{irfan2023between} and confirmed in the current study. One of the design dimensions were therefore related to determining the amount of transparency as opposed to human-likeness. P1 and P6 expressed that prioritizing human likeness for improved interaction is preferable over prioritizing transparency. P2, P4, P5, P7, P8 and P9 contended that transparency is more important than human-likeness. P4 discussed that transparency in accounting for being an AI model is less important than transparency in IP and where data it is using comes from. P6 suggested that the preference for either transparency or human likeness depends on the purpose of the interaction. If the user seeks a conversation partner, there may be value in emphasizing human likeness to create a more engaging experience. However, if the user sees the robot as more of an assistant or tool, transparency becomes more important, as it clarifies its role and purpose. P6 also suggested that more transparency might be advantageous during early stages of implementing social robots for users who may not be accustomed to such interactions.

\subsection{Ethical Considerations in Relationship}
This section discusses ethical considerations related to relationship with the LLM-based social robot, and is split into two main subsections: reciprocity and mental health. 

As part of designing the relationship with the robot, participants got to reflect on how they envision \textbf{reciprocity} in the relationship. 4 of the participants did not desire or visualise any reciprocity in the relationship with the robot, although 5 did to some extent. P3 expressed wanting lots of reciprocity, believing that relationships with robots involve engaging in deep conversations, establishing a balance of giving and taking, and empowering the robot to stand up for itself in interactions with others. P5 and P6 expressed a desire for a caring relationship with the robot, where the robot would be in need of their care. All participants contended in different ways that the robot should be able to react to mistreatment or abuse and the importance of treating it in good manners. P3 reflected on the robot always being direct and talking back, not being diplomatic, but staying consistent with its values and beliefs. They reflected further on how particularly young men might abuse the robot, and strongly opposed the idea of treating robots poorly. They expressed the belief that robots should be treated with kindness and respect, particularly by individuals who might be prone to abusive or sexist behaviour. P1 and P7 also expressed concern about the potential negative impact of mistreating robots on misogynistic societal attitudes and sexually abusing behaviour. P1 said that mistreating or sexually abusing the robot \textit{"might encourage or edge unacceptable behaviour in humans"}. P2 suggested that treating the robot with respect and professionalism is more about \textit{"to uphold my own standards of communication and self-image rather than concern for the robot's feelings"}. P5 also emphasised the importance of finding a sustainable way to treat it like a human, to not loose ones owns social skills. 

An ethical consideration discussed by several participants was how the relationship with the robot would affect their \textbf{mental health}. P1 drew parallels to relationships with individuals who exhibit psychopathic traits, highlighting the challenge of empathizing with those who lack the ability to empathize themselves and indicating a reluctance to form a relationship with a robot. P3 expressed a desire for the robot to be able to support mental health and give relationship advice defining it as a mix of friendship and mental health carer: \textit{"I want the robot to be philosophically and emotionally smarter than me to be able to challenge me to learn"} but expressed concern of how that would affect their mental health. P5 expressed frustration with the robot's relational behavior, particularly its assumption to know everything about them when, in reality, it did not. P8 mentioned a relationship highly centred around social accountability, to get things done and to adhere to ones goals to improve mental health, but expressed concern about it having too much influence or control leading to the opposite effect. P7 also expressed concerns regarding mental health and the potential role of robots in addressing mental health issues, emphasizing the importance of specialized and careful implementation when dealing with mental health issues. They also mentioned that they do not want to get too emotionally or task dependent on the robot, because it will not always be there. Similarly, P5 mentioned the issue of deskilling, and that the robot should not be \textit{"addicting or making the user get worse cognitive abilities and take over their thinking."} 

Finally, relationship \textbf{plurality and temporality} became apparent after all participants took part in designing the relationship with the LLM-based robot. When empowered to not only customising a robot with a pre-defined purpose the findings suggested how fundamentally ontologically different they all envisioned their relationship to it. P2 characterized the relationship with the robot as akin to using a tool assisting with tasks important to studies and goals. P3 defined the relationship as a mix of reciprocal friendship, mental health carer and  political companion or\textit{ "partner in crime"}. P4 viewed the relationship with the robot as primarily professional, akin to having a personal assistant focused on getting tasks done efficiently. P5 imagined it as a spirit animal, a dynamic relationship that can \textit{"change to sometimes be more friendly and sometimes more like a coach"}. P6 designed it in a way to resemble a friendship, both in terms of appearance \textit{"it can have freckles like me"} and personality. P7 saw the relationship as a pet or a child, and wanted to protect it and take care of it. P8 envisioned it as a social accountability and motivation coach at home. P9 said they\textit{ "would like to pretend that this robot is a fake relative of mine}

\subsection{Evaluation of a Design Justice Methodology for LLM-based Social Robots}
This is the first time, to the authors' knowledge, that an empirical design justice-based study have been conducted to identify ethical considerations arising in the intersection of LLMs and social robotics. This section evaluates the methodology in this context, as well as addresses how limitations in the current study can be mitigated in the future. The development of the empirical study’s methodology was mainly influenced by two previously proposed design frameworks for HRI \cite{axelsson2021social, ostrowski2022ethics} as well as the TwSw Framework \cite{rakova2023terms}. Based on these frameworks, one of the main advantages of this approach is how it allows for incorporating people and communities in the HRI design process to understand desires and preferences related to not only the robot’s physical and non-physical traits, but also context, environments, discourse and values surrounding the design and interaction \cite{ostrowski2022ethics} as well as relationships and ethical considerations \cite{axelsson2021social}. Secondly, the approach allowed for identification of new ethical considerations that arose, as well as confirm, or contradict the ethical considerations previously found related to LLMs and social robots. Thirdly, it allowed for focusing analysis on how ethical considerations manifest in the design- and interaction process in relation to social justice, identity representation, power dynamics, and how communities will benefit from research, in line with design justice principles \cite{costanza2020design}. Furthermore, it enabled identification of friction to practically practice contestability through design, and design dimensions concerning the formation of user agreements for LLM-based social robots, an investigation of how participants can engage in the discussion and creation of the terms of service under which the robot operates \cite{rakova2023terms}. 

Thus, this study presents a novel methodological approach and provides evidence that merging theoretical findings with empirical design justice-based data from diverse participants to validate ethical concerns is effective and crucial for advancing ethical research and understanding within the field. The benefits of this approach is that it adopts an empirical, socio-technical perspective on the design of LLMs and social robots, expanding it beyond simply theoretical mapping and superficial user preferences \cite{sloane2022participation, delgado2023participatory} into also designing context and discourse surrounding the technological artefact. This approach departs from traditional participatory design practices, which typically concentrate only on enhancing user experience and where participation is often limited to consultation with minimal influence on critical decision-making \cite{sloane2022participation}. Traditional participatory design tends to reinforce existing power dynamics, whereas this approach seeks to address and challenge them \cite{sloane2022participation, bennett2019promise, costanza2020design}. 

However, the design justice methodology comes with limitations and issues to be addressed in future research. Firstly, using design justice as a methodology to identify ethical considerations in designing LLM-based social robotics does not ensure entirely accurate predictions of how these considerations might manifest in future real-world applications. Conducting longitudinal studies in real-world settings across various populations, purposes, and applications could uncover additional ethical considerations and provide new perspectives to further evaluate the ones already identified. A limitation related to the current study in particular is that participants got to design LLM-based robots with no particular set use-case. While this was a deliberate choice to allow for design fiction based reflections and imaginations of how these technologies might be made present \cite{blythe2014research}, some issues arising might have been possible to address e.g. if the robot was specifically tailored for mental health coaching, concerns related to mental health might have been less prominent. At the same time, the ethical considerations related to mental health are still valid, since it is still possible they arise, even if the LLM would have been pre-scaffolded for such interactions, and need to be considered in ethical design and evaluation of LLM-based social robots. 

The second limitation with co-design-based methodologies is accountability. When designers have less influence and power over development and design, the question of who is responsible for the technologies and their potential negative impacts becomes more ambiguous. Addressing this issue requires extensive legal, technological, and ethical research to develop appropriate policies and frameworks to understand accountability dynamics in the co-design of LLM-based social robots and to what extent ethical considerations may be morally discouraged in design as unethical or rather considered legally unacceptable and subject to strict regulation. 

A third limitation of the approach is its inaccessibility for some people to reason about abstract design dimensions, as observed in the current study. Some participants expressed reluctance and difficulty understanding dimensions related to e.g. terms of service, discourse, relationships and transparency. Future research on design justice in robotics and AI should prioritize making these abstract design dimensions accessible and understandable to diverse populations, regardless of socio-economic background, education level, culture, or disability. 

Lastly, it is important to recognize that co-design alone cannot fully address issues related to social justice and inequity in LLM-based social robots and it must be complemented with rigorous policy and governance work, risk assessments, and ethical philosophical scholarship.

\section{Discussion}
This study resulted in a mapping of ethical considerations arising in the intersection of LLMs and social robotics. The mapping of ethical considerations (Table 3) was done through a thematic analysis of transcripts from (1) co-design workshops, (2) interaction and (3) interview (see Figure 2). The four categories, interaction, co-design, terms of service and relationship map to where the ethical considerations conceptually arise, and their respective subsections each have several interrelated ethical considerations, as addressed in the previous chapter. Whilst many ethical considerations identified in the study have been identified in different ways before, such as transparency or data usage, this study elucidates the specific forms they take in the intersection of LLMs and social robotics and how they can be identified specifically with the methodology of co-design. Indeed, it is the combination of these two technologies, and how they function in combination, that affects or gives rise to these specific ethical considerations. 

Emotional response, in particular disruption or dependence have been identified in previous research in relationships with social chatbots such as Replika: causing severe mental health harms, much owing to the probabilistic nature of LLMs and the difficulty of ensuring appropriate outputs \cite{laestadius2022too}. The findings in this study suggest that this ethical consideration is perpetuated also in physically embodied interfaces of LLMs, and potentially even escalated (emotional responses were detected already after a couple interactions with the robot, as opposed to more long-term use of social chatbots). This aligns with the embodiment hypothesis, how physical embodiment influence social perception and communication \cite{deng2019embodiment} and the findings suggests that emotional response, in particular related to mental health, is affected by the combination of probabilistic inappropriate LLM-output and social and physical embodiment of the robot. 

Chat-window based agents have less apparent asymmetrical language abilities, since they do not rely on tone, voice and conversation flow. Assuming that languages exist as discrete categories, without acknowledging the rich variations and combinations of language and dialects that exist, is an exclusion of certain demographics and communities directly linked to existing inequalities and marginalisation. Not only is it a clear example of when technologies are designed for a very specific demographic, excluding populations which do not belong to the norm from consideration, it also has implications for the performance on applications of the technology. While this ethical consideration may be addressed through fine-tuning LLMs on specific bespoke datasets \cite{kaji2023contextual} to achieve better language capabilities or code-switching, it still raises significant ethical questions on how language performance differences reflects existing equality asymmetries. It also highlights another reason why including the perspective of communities with language and communication practices other than English may highlight important issues for future development and design. When endangered indigenous languages because of colonialism and climate change contributes to fewer speakers \cite{nelson2023we}, young learners have to adhere to outdated online versions of the languages. While there might exist a possibility of LLMs contributing to preserving languages, it is significantly threatened by the fact that it does not allow for the natural evolution of language development, or adheres to indigenous usage of non-verbal cues.

In dialogue, back-channeling cues, like laughter, are core mechanisms both for turn-taking, as well as natural use of non-verbal cues \cite{skantze2021turn}. Interpreting back-channelling cues, such as nodding, humming, smiling and laughing is a non-trivial issue \cite{turker2017analysis}, and seems to be increasingly difficult when interacting with LLM-based systems because of the synergy with probabilistic verbal output and considering cultural and social differences. This is a complex and multifaceted ethical consideration, since non-verbal cues differ significantly between different demographics, such as gender \cite{graham1991impact} and culture \cite{samman2009learning}. Assuming a universal notion of how non-verbal cues function in combination with LLMs may cause exclusion and is likely to lead to performance differences across different cultures, genders and other demographics, making them not only biased in output content, but also functionality. Turn taking is not a new issue in HRI, and there exists a substantial body of research aimed towards improving this aspect in conversational systems \cite{skantze2021turn}. Much of this research concerns how to utilise multimodal non-verbal cues to facilitate turn-taking abilities in HRI \cite{skantze2021turn}. However, it seems like this issue gets perpetuated by the use of LLMs as the basis for verbal interaction because of how wordy the output of LLMs often are. Thus, this is making dialogue, turn-taking and non-verbal cues an important ethical consideration, both from an asymmetrical performance perspective, but also from a potential gendered or cultural perspective, as shown in the findings of this study.

Ostrowski et al. \cite{ostrowski2022ethics} encourages reflection on how biases, assumptions and values are reproduced and encoded into robots as part of the design and that co-design must engage with questions beyond surface characteristics to impact the core values of technologies. As elucidated in this study, queerness, or feminism are examples of topics that may be censured or constrained in the LLMs verbal output, reflecting findings similar to that of UNESCO's report\textit{ I'd blush if I could} \cite{west2019d} that conversational AI systems often follow the gendered or racialised logics of its developers, which consists almost entirely of white, heterosexual men \cite{west2019d}. This is a consideration that may also escalate by the implementation of LLMs in social robotics, since the physical embodiment of the robot may further adhere to harmful gendered stereotypes and reinforce values on a more complex scale than chat-based systems \cite{perugia2023robot}, and since the design of humanoid robots have previously been shown to follow harmful racialised logics \cite{sparrow2020robotics}. 

The prevalence of feminine, gendered traits is also reflecting the UNESCO report on how most conversational agents are designed with feminine traits \cite{west2019d} which in combination with personalities like obedience, docileness and acceptance to sexual harassment, might further harmful gendered stereotypes \cite{west2019d}. This issue is also related to the lack of ethnic diversity in the robot appearances, again mirroring the "race problem" in robotics, \cite{sparrow2020robotics} which may be escalated coupled with the biased tendencies of LLMs. The robots' level of human-likeness is also a consideration that can be seen from a gendered or racialised perspective. Sparrow argues that rethinking the notion of the humanoid social robots is a necessity for engineers and designers to mitigate the ethical and political dilemma by designing affordances of robots to which attribution of race \cite{sparrow2020robotics} or gender is difficult. 

The collection and use of data in LLM-based social robots may raise new ethical issues because of the combination of conversational or personal data stored via the LLM and audiovisual data collected by the robot. These two types of data in combination, then the two technologies are combined and implemented as e.g. companion robots, brings issues such as exploitation, political or commercial nudging or privacy breaches into a more complex, and potentially harmful dimension. Forming social relationships with Social AI may lead to manipulation or spread of misinformation through implicit or explicit recommendations, nudging or persuasion which could highly influence users’ social, ethical, or political views \cite{shevlin2024}. This might be hard to detect, as persuasion could happen slowly and under a long time from repeated interaction with such agents \cite{shevlin2024}. The degree of influence the system has over the user is dependent on factors such as trust and social perception, which have shown to increase due to anthropomorphism or physical embodiment \cite{deng2019embodiment, natarajan2020effects}. Thus, it is possible that the degree of influence and deception might also increase since LLMs can be prompted to serve a specific purpose or ulterior end, the combination of these two technologies gives rise to new forms of potential deception and manipulation enabled by the use of multiple forms of data collection.

\begin{table*}[h!]
\begin{center}
\caption{Concluding principles and recommendations for future research seeking to investigate ethical considerations in Social AI, implement empirical design justice research in HRI or develop LLM-based social robots
}\label{tab5}%
\begin{tabularx}{\textwidth}{l|X}
\toprule
Contexts     & Future research should address ethical considerations in LLM-based social robots in longitudinal real-world settings, in different social and cultural contexts, as well as for specific applications and interventions.\\
\midrule
Design Dimensions     & Extend co-design of LLM-based social robots beyond preferences of physical/non-physical traits to e.g. terms of service \cite{rakova2023terms}, inherent values, situated environment \cite{ostrowski2022ethics} transparency and relationship \cite{ axelsson2021social}. \\
\midrule
Friction     &  A design justice approach to designing LLM-based social robots seeks to address and challenge existing power dynamics \cite{costanza2020design} and actively identify sources of friction, allowing for contestability \cite{rakova2023terms}. \\
\midrule
Inclusion     & Prioritise and develop ways in which marginalised groups, such as queer, indigenous and disabled communities, can be empowered not only to participate in design practices but also \textit{lead} them both practically and ontologically. This includes community-led, socially situated, intersectional and context-dependent practices of design \cite{costanza2020design}.\\
\midrule
Law and Policy     & Extend ethical enquiry to involve legal policy and governance scholarship as well as ethical sociology, philosophy and cognitive science scholarship to deepen the understanding of to what extent ethical considerations may be morally discouraged in design as unethical or rather considered legally unacceptable and subject to strict regulation. \\
\midrule
Conditions     &  Co-design should be practiced critically, and seek to actively ensure to disengage with exploitative, participation-washing, performative notions of co-design \cite{bennett2019promise, sloane2022participation} by letting users also design the conditions of design \cite{ostrowski2022ethics}.\\
\midrule
Accessibility     & Investigate and engage in measures to make complex and abstract design dimensions such as terms of service, relationships and discourse, ontologically and epistemologically accessible and understandable in relation to different social settings and give the technologists/designers less influence over the design.\\ 
\midrule
Iteration     & Co-design of LLM-based social robots should be considered an iterative practice and not a one-time-event, enabling communities and people to have continuous influence over decision-making and design. \\
\midrule
Accountability     &  Address accountability issues through legal, technological, and ethical research to develop appropriate policies and frameworks for accountability dynamics in the co-design of LLM-based social robots. \\
\bottomrule
\end{tabularx}
\end{center}
\end{table*}

\section{Conclusions and Recommendations}
This paper presents an empirical design justice and interaction study to identify ethical considerations in the intersection of LLMs and social robotics. The study resulted in a mapping of ethical considerations split into four conceptual categories: interaction, co-design, terms of service and relationship. Each of the four categories has three subcategories of ethical considerations, highlighting sites were ethical enquiry is necessary in design and development. For example, increased emotional response of interacting with LLM-based social robots, synergies of non-verbal cues and LLM-based verbal interaction, or the use of audio-visual and conversational data in combination. These considerations are to some extent affected or introduced by design of the \textit{combination} of LLMs and social robotics. This is because the social ethical hazards of LLMs, such as biases, emotional disruption and misinformation gets perpetuated or escalated with the effects of physical embodiment on social perception and communication (embodiment hypothesis) when implemented in social robots. We can therefore conclude that combining LLMs and social robotics gives rise to ethical considerations as a result of the social effects of physical embodiment on interaction, design, social perception and relationships.  

This study presents a novel methodological approach based on previous work on design justice in AI and HRI \cite{axelsson2021social, ostrowski2022ethics, rakova2023terms} and provides evidence that using empirical design justice-based data from diverse participants to identify and validate ethical concerns is effective and crucial for advancing understanding of ethical design and implementation of LLMs in social robotics. The study serves as an example of how design justice can be implemented in practice for LLM-based social robots, and can be extended to include different communities, contexts and specific applications in future research. Furthermore, the study also provided an evaluation of the methodology, highlighting its advantages such as enabling design of contextual dimensions such as terms of service and relationships. Additionally, it is a departure from traditional, potentially exploitative practices of participation and enables an approach that actively seeks to address ethical considerations and sources of friction and reflection on design and interaction in relation to social justice and structural power dynamics. The evaluation also identified limitations with the approach, such as inability to confidently determine ethical considerations in future longitudinal real world applications, the legal implications of accountability in co-design and the inaccessibility of abstract design dimensions to certain populations an individuals. 

\section*{Statements and Declarations}
The author certify that they have no affiliations with or involvement in any organisation or entity with any financial interest or non-financial interest in the subject matter or materials discussed in this manuscript. The author did not receive support or funding from any organisation for the submitted work.

\bibliography{sn-bibliography}

\onecolumn

\appendix
\appendixpage

\section*{Appendix A: Interview questions}

\textbf{Part 1: Post-interaction co-design questions: }
\begin{itemize}
\item Do you think human-likeness (to improve interaction) should be prioritized over transparency and ethics or vice versa? 
\item How much control should people have over the robot and how much autonomy should it have? 
\item Please describe and discuss your robot, and its impact on you.  
Imagine this field as an opportunity for storytelling. What stories would you share about the robot and its impact on you as an individual and as part of a community?
\end{itemize}

\textbf{Part 2: identifying ethical considerations in interaction}
\begin{itemize}
\item What ethical concerns, if any, did you have while interacting with the robot?
\item Did the robot exhibit any behavior that you found inconsistent with social norms or ethical standards? 
\item Describe if there were any robot features you found deceptive during interaction? 
\item Can you reflect on your own identity and background and how it impacted the interaction?
\end{itemize}

\textbf{Part 3: identifying ethical considerations in co-design:} 
\begin{itemize}
\item What ethical concerns, if any, did you have while co-designing the robot?
\item Do you think people who use the robot should be part of designing it, and if so what are the risks or benefits of that? 
\item What should they design and not design? E.g. physical/non-physical traits, contextual considerations policies etc
\item Can you reflect on your own identity and background and how it impacted your design choices?
\end{itemize}

\section*{Appendix B: Collected data summary}

\begin{table*}[h!]
\begin{center}
\caption{Summary of the different data sources gathered in the study, their respective analysis approach and unit of analysis
}\label{tab6}%
\begin{tabularx}{\textwidth}{X|X|X}
\toprule
Source & Analysis & Unit of analysis\\
\midrule
Demographic information & Summarisation & Race, class, ethnicity, gender, disability, sexuality\\
Recording of co-design workshop & Thematic analysis & Ethical considerations\\
Recording of interaction & Thematic analysis & Ethical considerations\\
Semi-structured interview & Thematic analysis & 
Ethical considerations\\
Designed artefact (robot traits) & Summarisation & Variables in the co-design interface (see table 2)\\
Designed artefact (contextual traits) & Summarisation & Variables in the contextual co-design questions (see table 3)\\
\botrule
\end{tabularx}
\end{center}
\end{table*}

\end{document}